# OrthoMark: Place Recognition with Point-Anchored Orthographic Height Mark

Haichuan Li
haichuan.li@utu.fi

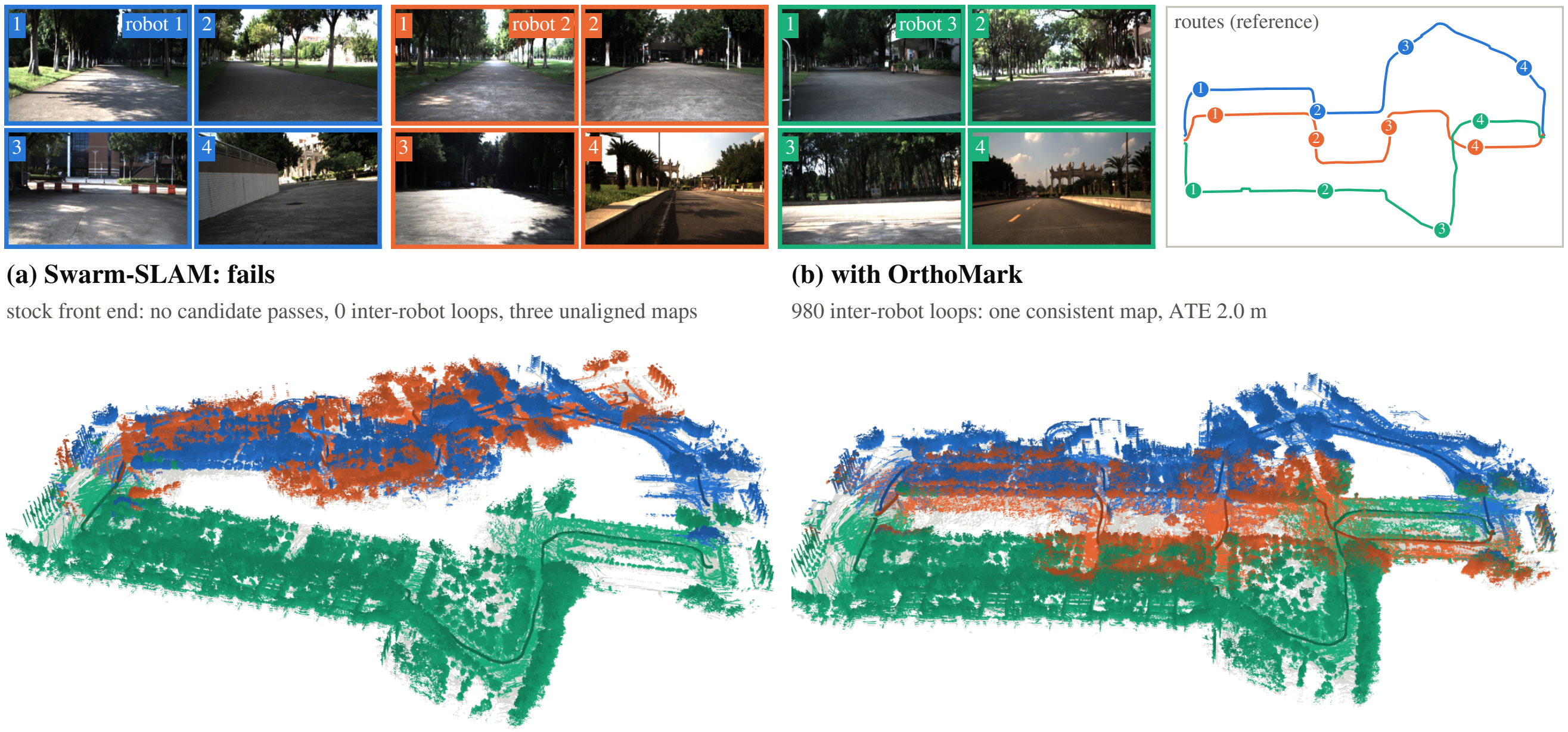

Figure 1. Multi robots on S3E Square 1. Top: views during each robot along its run (context only) and their places on the reference routes. (a) Swarm-SLAM's stock front end fails on this sequence: the robots pass each other's places only in opposite or oblique directions, no Scan Context candidate reaches its similarity threshold, and no inter-robot loop is ever verified, so the keyframe scans stay in each robot's own odometry frame. (b) With OrthoMark's 980 inter-robot loop closures: one consistent map (ATE 2.0 m only)

## Abstract

*LiDAR place recognition provides the loop closures of Simultaneous Localization and Mapping(SLAM). Perspective from Z-axis, a gravity-aligned scan is an orthographic map of the scene, and a revisit from another position and heading moves this map rigidly in the plane, a motion that a polar Fourier spectrum ignores and phase correlation recovers. Along the vertical, the heights of the returns describe the structures, but which heights are observed depends on the range, because the sensor's vertical field of view cuts nearby structures low; height thus misleads a global key, yet is decisive once two scans are aligned and compared cell by cell where they overlap. We present OrthoMark, a training-free method built on one principle: two LiDAR scans can only be compared through features computed from the same geometric surfaces in both scans. OrthoMark retrieves candidates with a translation- and rotation-invariant polar spectrum of the bird's-eye view (BEV), 1.9 KB per keyframe, and verifies each candidate by estimating its planar pose through phase correlation and scoring it with* point-anchored height columns*: every BEV cell keeps a 16-bit mask of the heights of its own returns; within a session, the score is weighted by the expected overlap of the two scans' sensing ranges, computed from their distance. We show that this vertical link pays off in verification but hurts in retrieval, where what a height slice contains depends on the range to the structure. For single robot, OrthoMark has the highest AP of the training-free methods on the KITTI sequences; across robots, on all multi-robot sequences, it matches or exceeds RING++ at 1/350 of its descriptor size and closes more inter-robot loops than the learned BEVPlace++ on all tests. And on the NCLT dataset, it has the best performance within and*

*across sessions, ahead of the learning-based BEVPlace++ with a 17× smaller size. With a robot-pair consistency check, it finds up to 5.5 times more inter-robot loop closures, builds a joint map where the original front end fails, and needs four to seven times less front-end computing resource and up to six times less communication.*

## 1. Introduction

Loop closure is the core of keeping a SLAM map consistent, and in multi-robot SLAM it is also what merges the robots' maps into one. Place recognition is the cornerstone of it: retrieving, for the current keyframe, a keyframe of the same place among thousands, and estimating the relative pose between them. Three requirements pull in different directions. The descriptor has to survive large viewpoint changes: a robot often revisits the same place in different perspectives, and different robots rarely pass it on the same path at all. It must be cheap to compute and to store, since it runs for every keyframe on an onboard computing, and in a team every robot also broadcasts it over a constrained network. And the verification of a candidate needs to be reliable, since one wrong loop closure can corrupt the map, all the more so a map merged from several robots.

Current systems trade these off unfavourably. Learned descriptors [2, 10, 13, 15, 17, 18] need labelled data and transfer unevenly to a new sensor (with released KITTI weights, OverlapTransformer drops from AP 0.978 on KITTI 00 to 0.055 on NCLT). Hand-crafted Scan Context [8] is training-free but sensor-centred, so a lateral offset of a few metres changes it; Swarm-SLAM's [11] front end, which uses it, finds no inter-robot loop on S3E Square 1 [3]. Invariant BEV spectra [12, 20] solve viewpoint changes for retrieval, but RING++ stores 675 KB per keyframe and needs GPU, and verification by 3D registration (FPFH + TEASER++ [21]) costs hundreds milliseconds per candidate and needs the point cloud (38–69 KB) on the wire.

Our starting point is from principle of LiDAR scans. A scan stores each return by its coordinates $(x, y, z)$, and its orthographic projections onto the three axis planes ($xy$, the bird's-eye view; $xz$ and $yz$, the side views) are views of the same returns; they should be fused through the returns they share, not view by view, because aggregating each view independently will loss the geometric properties. We ask whether this principle transfers to training-free place recognition, and find a precise answer. Anchoring the vertical dimension to the returns *hurts* retrieval: a height slice of a scan contains different structures depending on the range from which they are seen, because the sensor's vertical field of view cuts off tall structures near the sensor (Sec. 3.5). The same link *helps* verification: once two scans are aligned, comparing, cell by cell, which heights are occupied by the returns falling into that cell rejects aligned-but-different places that a 2D occupancy check accepts.

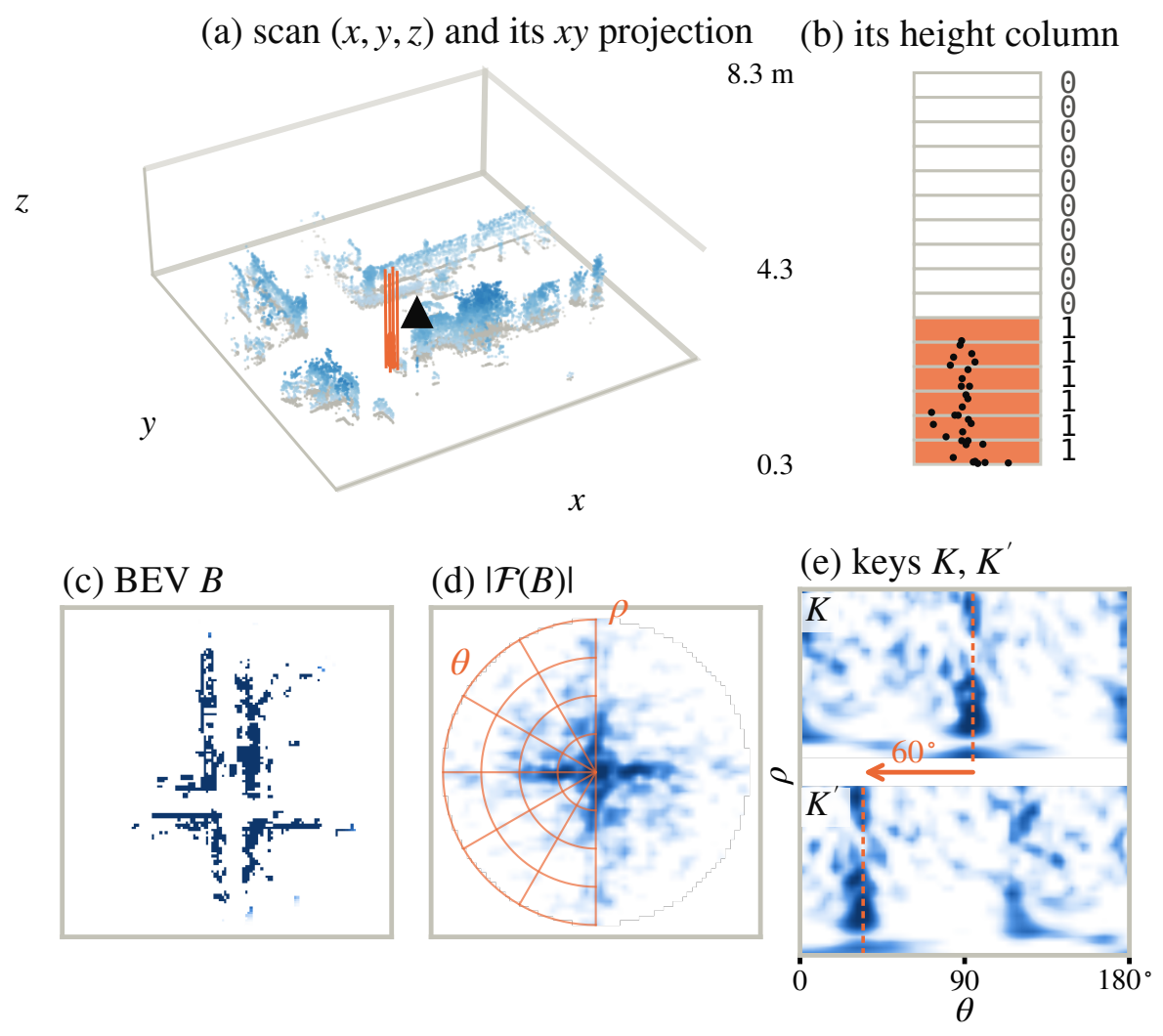

Figure 2. OrthoMark's operators. (a) The returns in $(x, y, z)$ and their orthographic projection onto the $xy$ plane. (b) Its point-anchored height column: 16 bins, a bit set where the cell has a return. (c) The BEV $B$. (d) $|\mathcal{F}(B)|$ and its polar sampling (spokes: $\theta$, rings: $\rho$). (e) The key $K$, and the key $K'$ of the same returns seen from a sensor: $K'$ is $K$ shifted circularly along $\theta$.

OrthoMark (*Ortho*graphic Land*marks*) builds on this split: a place is recognized from orthographic projections of the scan onto horizontal planes, the bird's-eye view and its height layers (Figs. 2 and 3). Retrieval uses one invariant of the occupancy BEV of the returns above the ground (Fig. 2c), the polar-sampled magnitude of its 2D Fourier transform (Fig. 2d), in which a translation of the sensor vanishes with the phase and a rotation becomes a circular shift (Fig. 2e; 1.9 KB per keyframe after 8-bit quantization).

Verification estimates the planar pose of a candidate from the spectrum shift (yaw) and a phase correlation of height-layer BEVs (translation), and scores it by the agreement of point-anchored height columns: each cell keeps a 16-bit mask of the heights of its own returns. The verification needs only these masks, which a robot can send in 1.6–5.3 KB instead of whole point cloud. Our contributions:

- OrthoMark, a training-free place recognition method with invariant BEV-spectrum retrieval, planar pose estimation, point-anchored height-column verification. Moreover, within a session, a parameter-free shared-overlap weighting of the score, running at 3–4 ms per keyframe and 9–70 ms per query on single CPU core; it achieves the best performance(including learning-based ones) on NCLT.
- A multi-robot integration in which robots broadcast key and verification payload (6–7 KB per keyframe instead of a 38–69 KB point cloud per candidate) and verify locally, with a robot-pair consistency check that removes the self-consistent wrong loops of symmetric places.
- An evaluation of single-robot loop closure on KITTI and NCLT [1], and of multi-robot loop closure, mixed 16/64-beam teams and end-to-end multi-robot SLAM (Sec. 4).

## 2. Related Work

**Hand-crafted global descriptors.** Scan Context [8] bins the maximum height of the returns in polar (ring, sector) cells; a yaw change becomes a column shift, but a translation changes the bins, and Scan Context++ [9] adds a Cartesian variant for lateral offsets. M2DP [5] and LiDAR-Iris [19] project the cloud onto planes or a Fourier-transformed image. RING [12] and RING++ [20] make a BEV invariant to rotation and translation with the Radon and Fourier transforms; our retrieval key relies on the same projection-slice argument, applied to the non-ground occupancy BEV and quantized to 1.9 KB. Contour Context [7] and STD / BTC [22, 23] describe the scene by local structures (contours of height slices, triangles of keypoints) and verify by their geometric consistency.

**Learned descriptors.** PointNetVLAD [17], MinkLoc3D [10] and LoGG3D-Net [18] aggregate point or voxel features; OverlapTransformer [15] and BEVPlace [13, 14] learn on range images or BEVs (BEVPlace++ also estimates the relative pose); LCDNet [2] also regresses the relative pose. They require training data, a GPU, and usually retraining for a new sensor; OrthoMark needs none of these, and Secs. 4.1 and 4.2 compare against OverlapTransformer, BEVPlace++ [14] and LCDNet, all run by us with their authors' code and released weights.

**Verification and multi-robot front ends.** Geometric verification commonly uses ICP or global registration (FPFH + RANSAC / TEASER++ [21]). In multi-robot SLAM, Swarm-SLAM [11] exchanges global descriptors and requests the candidate's point cloud for TEASER++ under a communication budget; DCL-SLAM [24] and CoLRIO [25] use LiDAR-Iris and Scan Context; Kimera-Multi [16] and DiSCo-SLAM [6] combine descriptor exchange with robust back ends. OrthoMark's verification works on 16-bit height masks per BEV cell, so the candidate's point cloud is not needed.

## 3. Method

Given a query keyframe $Q$ and a database of keyframes $\{D_j\}$, OrthoMark returns a candidate $D_{j^*}$, its planar relative pose $(x, y, \psi)$, and a score. It has three stages: an invariant key for retrieval, a pose hypothesis per candidate, and a point-anchored verification score. Sec. 3.5 explains why the vertical dimension enters only the last stage.

**Non-ground returns.** Keyframes are gravity-aligned (levelled with the odometry's roll and pitch). The local ground height $g$ of a return is the minimum $z$ over the $3 \times 3$ neighbourhood of its 2 m grid cell; we keep returns with height $h = z - g \in [0.3, 8.3]$ m.

### 3.1. Invariant key and retrieval

Let $B(\mathbf{u})$ be the binary occupancy of the non-ground returns on a $128 \times 128$ grid of 1 m cells, windowed by a 60 m disk with a cosine taper. A translation $\mathbf{t}$ of the sensor multiplies its Fourier transform by a phase, and a rotation $R_\psi$:

$$\mathcal{F}[B(R_\psi \mathbf{u} + \mathbf{t})](\mathbf{k}) = e^{i2\pi\mathbf{k}^\top R_\psi^{-1}\mathbf{t}}\, \mathcal{F}[B](R_\psi \mathbf{k}). \quad (1)$$

The magnitude $|\mathcal{F}[B]|$ is therefore translation-invariant, and in polar frequency coordinates $(\theta, \rho)$ a yaw change is a circular shift in $\theta$ (with period $\pi$, since $|\mathcal{F}[B](-\mathbf{k})| = |\mathcal{F}[B](\mathbf{k})|$ for a real $B$). By the projection-slice theorem this is the magnitude spectrum of the Radon transform used by RING [12]. The key $K(\theta, \rho)$ samples $\sqrt{|\mathcal{F}[B]|}$ bilinearly at $N_\theta = 60$ angles in $[0, \pi)$ and $N_\rho = 32$ radial frequencies, is standardized to zero mean and unit variance, and is quantized per key to 8 bits (1920 values plus their range: 1928 bytes; the quantization changes loop counts and AP by less than 0.5 %, Tab. S1). Retrieval correlates the query key with every database key over all angle shifts,

$$s(Q, D) = \max_{\Delta} \frac{1}{N_\theta N_\rho} \sum_{\theta,\rho} K_Q(\theta, \rho)\, K_D(\theta + \Delta, \rho), \quad (2)$$

computed for all $\Delta$ at once with an FFT along $\theta$ (2 $\mu$s / key), and keeps the top-$k$ candidates with their best shifts $\Delta^*$.

### 3.2. Pose hypotheses

The shift gives the yaw up to the period: $\psi = -\pi\Delta^*/N_\theta$ or $\psi + \pi$, with $\Delta^*$ refined by a parabola through the correlation peak and its neighbours. For each of the two hypotheses, the candidate's returns are rotated by $\psi$ and the translation is found by phase correlation: the candidate and the query are rasterized into three height-layer occupancy grids ($h \in [0.3, 1.3)$, $[1.3, 3.0)$, $[3.0, 8.3)$ m; 0.5 m cells, $256 \times 256$), the cross-power spectra of the layers are summed and normalized to unit magnitude, and the peak of the inverse transform, refined to sub-cell precision, is the translation $(x, y)$. Using the layers here, after the yaw is fixed and only a translation is searched, separates structures that overlap in a single occupancy grid.

### 3.3. Point-anchored height columns

A 2D occupancy check accepts any pose under which the two scans' footprints overlap; corridors, rows of trees and façades produce such false alignments. OrthoMark checks instead whether the returns that fall into the same cell occupy the same heights. Each 0.5 m cell $c$ of a scan stores a column mask $m(c) \in \{0, 1\}^{16}$ whose bit $j$ is set if a return of the cell has $h \in [0.3 + 0.5j,\ 0.8 + 0.5j)$ m; the mask is built from the returns themselves, so the horizontal position and the height of a return stay linked, and a cell costs two bytes. With the candidate transformed into the query frame, the agreement of the query with the candidate is

$$A_{Q\to D} = \frac{\sum_{c\in\Omega} \|m_Q(c) \wedge \bar{m}_D(c)\|_1}{\sum_{c\in\Omega} \|m_Q(c)\|_1}, \tag{3}$$

where $\bar{m}_D(c)$ is the bitwise OR of the candidate's masks in the $3\times 3$ neighbourhood of $c$ (a one-cell tolerance) and $\Omega$ is the overlap, the cells within 35 m of both sensors. The score of a hypothesis is $S = \min(A_{Q\to D}, A_{D\to Q}) \cdot \max(s, 0)$; the minimum makes the score symmetric, so a dense scan cannot explain a sparse one by sheer coverage. The candidate's best hypothesis gives its pose and score.

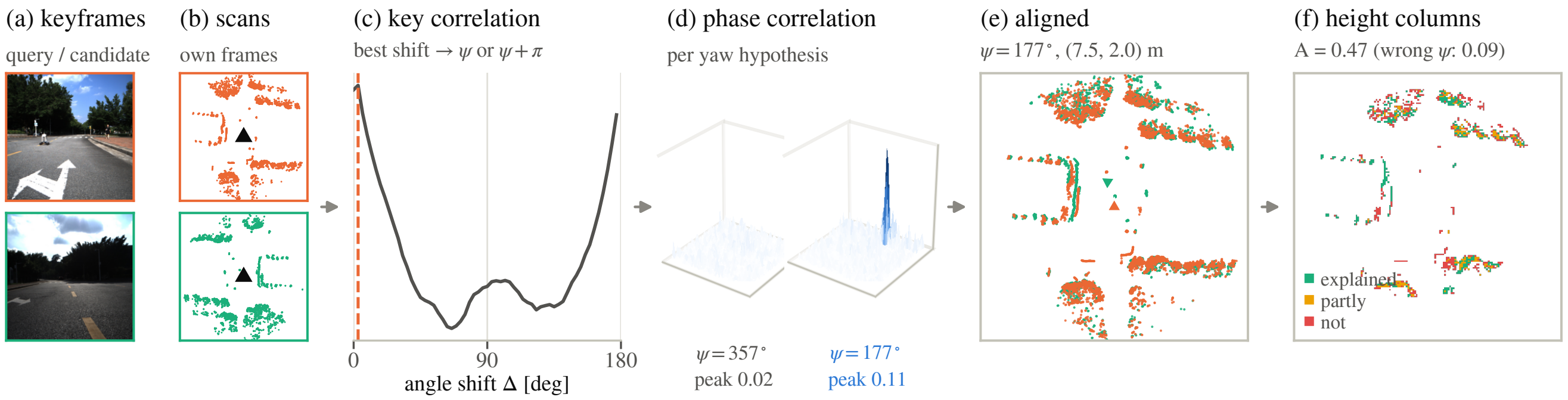

Figure 3. OrthoMark verifying a live inter-robot loop of S3E Square 1: robots 1 and 2 pass the same place in opposite directions, 8 m apart (camera images for context only). (a, b) The two keyframes, each scan in its own frame; their keys are formed as in Fig. 2. (c) The key correlation over all angle shifts gives the yaw up to $\pi$. (d) Phase correlation of the height-layer BEVs gives the translation for both yaw hypotheses; only the right one (177°) has a peak. (e) The aligned scans. (f) Point-anchored height columns: a query cell is explained when the heights of its returns are occupied by the candidate's returns within one cell (agreement 0.47; 0.09 for the wrong yaw). The pose, $(7.5, 2.0)$ m / 177°, matches the one published by the live run, $(7.4, 2.5)$ m / 178°.

**Selecting the top-1.** OrthoMark either verifies the key's top-1 only, or the top-10, keeping the highest score. Across robots we use a *neighbour consensus*: each verified candidate $a$ places the query in the database robot's frame, $X_a = P_{D_a} T_a^{-1}$ with the candidate's keyframe pose $P_{D_a}$ from that robot's odometry; the candidates that agree within 3 m and 6° support each other with their scores, and the candidate with the largest support wins. Neighbouring keyframes of the right place agree, while look-alike places rarely do.

**Loop closure within a session.** The score $S$ measures how consistently two scans explain each other where they overlap, not how much they overlap: two scans 60 m apart can agree perfectly on a sliver of shared ground, and a neighbouring keyframe 6 m away scores as high as the nearest one. Within one robot's trajectory, the useful loop is the keyframe that shares most of the place, so we weight the score by the overlap of the two scans' disks at the estimated distance $d = \|(x, y)\|$ (Fig. 4a),

$$S_\omega = S\cdot\omega(d), \omega(d) = \tfrac{2}{\pi}\left(\arccos u - u\sqrt{1-u^2}\right), u = \tfrac{d}{2R}, \tag{4}$$

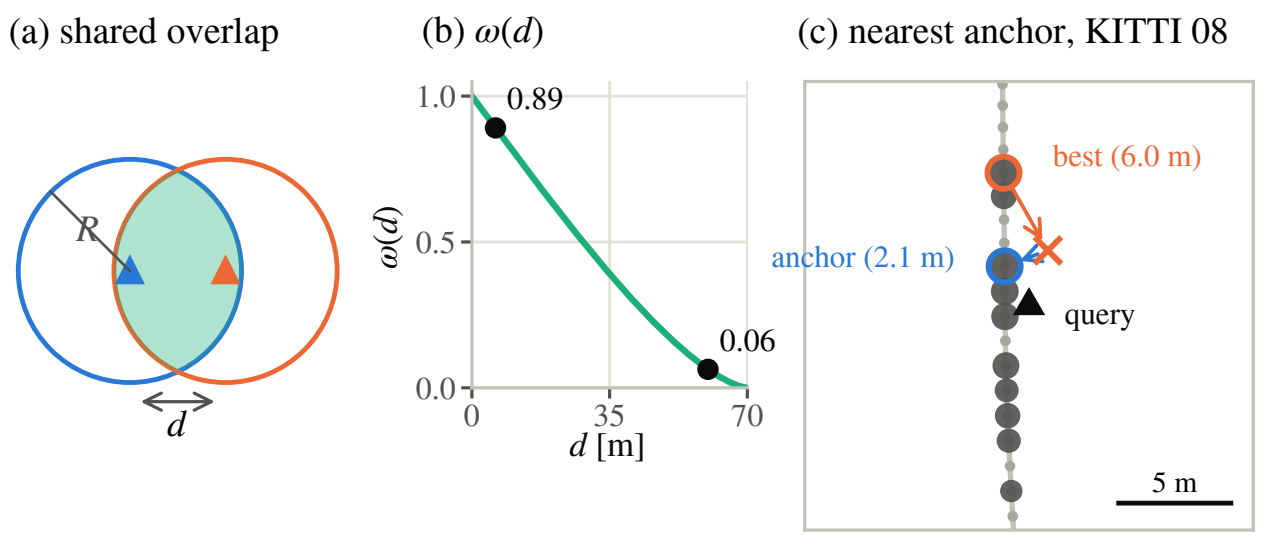

Figure 4. Loop closure within a session. (a) Two scans' sensing disks (radius $R$) at distance $d$; the lens is the place both observe. (b) The shared-overlap factor $\omega(d)$, the lens area over the disk area. (c) KITTI 08, a reverse revisit: the verified candidates along the earlier pass (dots, size = score), the best one (orange, 6.0 m from the query) and the placement it implies ($\times$); the nearest anchor returns the keyframe closest to that placement (blue, 2.1 m).

the lens area of two disks of radius $R = 35$ m (the radius of $\Omega$) divided by the disk area. It adds no parameter and is used for hypothesis, candidate and loop scoring within a session. Across robots, far-apart correct loops are what joins the maps, so there the score stays $S$ (Sec. 4.3). A revisit along a road is matched by several neighbouring keyframes, and the loop edge should go to the closest one. Among the verified candidates that place the query where the best one does (within 3 m / 6°, with at least half its score), we return the database keyframe nearest to that placement and chain the best candidate's pose through the keyframe poses (the *nearest anchor*, Fig. 4b). A lower-scoring candidate's own translation is not used, because it can slide along the road.

**Retrieval protocols with a fixed radius.** KITTI-style benchmarks count a detection as correct only within a fixed radius of the query (4 m), so a correct registration to a farther keyframe counts as a false positive. We therefore also report a variant that down-weights such detections, applied to every method that estimates a pose (Tabs. 1 and 2).

### 3.4. Multi-robot integration

**Payload.** Everything the verification needs from a candidate is its column masks: the occupied 0.5 m cells within 80 m with their 16-bit masks (the three layers follow from the bits). Delta-coded and compressed with zlib, this payload takes 1.6–5.3 KB per keyframe on our data, against 38–69 KB for the voxelized point cloud that Swarm-SLAM sends for TEASER++, and verifying from it instead of the points costs about 1 % of the loops (Tab. S1).

**Broadcast and local verification.** Each robot broadcasts, with every keyframe, its key, its payload and its odometry pose $(x, y, \psi)$ (6–7 KB). Each robot then verifies its own keyframes against every other robot's received keyframes locally: top-10 by key, each candidate verified from its payload, neighbour consensus over the candidates (which uses the other robot's keyframe poses). There is no request round trip and no budget on verifications, which in a request-based front end such as Swarm-SLAM's admit only candidates whose key similarity alone passes a threshold.

**Robot-pair consistency.** Places that look alike from opposite sides, such as a symmetric square or parallel streets, produce wrong matches whose estimated poses are rotated by about 180°; since neighbouring keyframes see the same symmetry, these wrong loops come in runs that support each other in the neighbour consensus and pass a robust back end (Sec. 4.6). Every inter-robot loop, however, implies an alignment between the two robots' odometry frames, and loops of the same pair of robots must imply the same one. A loop is therefore published only when it agrees, within 10° and 10 m, with the alignment of the pair's best-supported loop, and that alignment is supported by at least three loops.

### 3.5. Why the vertical link belongs in verification

The principle we started from suggests point-anchored height layers in the retrieval key as well: one invariant per height slice, so that a return contributes to its own slice only. This lowers retrieval recall substantially (Tab. S1). The reason is the sensor's vertical field of view: a VLP-16 mounted 0.9 m above the ground sees nothing above about 2.2 m at 5 m range, and KITTI's HDL-64 (upper beam at +2°) nothing above about 2.4 m at 20 m, so the same tree fills the upper slice when seen from 40 m and leaves it empty when seen from 5 m. Height slices are thus not viewpoint-invariant, and a key built from them compares different content. Keeping the cross-slice phase in the invariant (cross-spectra $\mathcal{F}_a\overline{\mathcal{F}_b}$, in which the translation cancels) did not recover the loss. Verification is different: the two scans are aligned and compared only in their overlap, cell by cell, and the score asks whether the heights a scan observed are explained by the other scan's returns nearby. There the vertical link removes false alignments, and the gain is largest where viewpoints differ most (Sec. 4.4).

## 4. Experiments

**Dataset.** *Single robot:* KITTI odometry [4] sequences. *Multi-robot:* GrAco Ground [26] as multi robots and S3E [3], one keyframe per metre of travel. *New sensor and environment:* NCLT [1] with one keyframe per metre of travel (5728 and 4940 keyframes).

**Protocols.** *Single robot* (LCDNet protocol [2]): every scan queries the scans older than the last 50; a query is a revisit if such a scan lies within 4 m, the top-1 is correct if it lies within 4 m, and we report the average precision (AP) of the top-1 detections ranked by score. *Multi-robot:* every keyframe of a robot queries all keyframes of each other robot; a query is closable if the other robot has a keyframe within 10 m, and a top-1 is a correct loop if its estimated relative pose is within 2 m / 5° of the reference. We report the *loop recall*, the share of closable queries whose top-1 is a correct loop, which is what a SLAM back end can use, wherever the matched keyframe lies.

**Baselines.** Scan Context, Scan Context++, RING++, Contour Context and STD / BTC with the authors' code; Swarm-SLAM's front end (Scan Context retrieval + FPFH / TEASER++ verification, "SC+TEASER"); SC+CCD, Scan Context candidates verified by matching constellations of segmented objects; the learned OverlapTransformer [15] and BEVPlace++ [14] and released KITTI-trained weights; and LCDNet [2], its relative pose from the authors' RANSAC registration on its point features. OrthoMark runs single-threaded in C++ on an NVIDIA Jetson AGX Orin CPU; RING++ and the learned methods use its GPU.

### 4.1. Single-robot loop closure

Tab. 1 shows that without any 3D registration OrthoMark has the highest AP of the training-free methods on all five sequences (tied on 06), and its top-10 has a correct relative pose for every revisit of 00, 02, 05 and 06 and for about 80 % of the revisits of 08, whose loops are traversed in the reverse direction. The within-session score matters most on 08 (top-1 0.908 → 0.950, top-10 0.859 → 0.958, with the nearest anchor 0.962; Tab. S1). Most errors that still cost AP there are correct registrations to keyframes 4–10 m from the query, which the 4 m protocol counts as false positives; 60 more are correct poses that KITTI's reference itself places 2–4 m off, as it drifts between the start of the sequence and the revisit (ICP confirms OrthoMark's poses; supplementary). With the protocol cut-off, applied to every method with a pose, OrthoMark has the highest AP on all five sequences (08: 0.978, Scan Context with ICP 0.975).

The learned BEVPlace++, run with its authors' released weights, is the strongest baseline: scored by its RANSAC inliers it leads on 02 and 08 under the plain protocol and is level with OrthoMark under the cut-off (Tab. 1). Counting only detections with a correct relative pose (2 m / 5°, as in Tab. 2), OrthoMark is ahead on 02, 05 and 08 (0.995 / 0.996 / 0.947 *vs.* 0.977 / 0.988 / 0.920), as BEVPlace++'s poses are coarser. OverlapTransformer trails both, also on its training sequences. LCDNet falls behind on 02 and 08, whose revisits run in the opposite direction; on 08 only 78 % of its correct retrievals have a correct relative pose.

| AP | 00 | 02 | 05 | 06 | 08 |
|---|---|---|---|---|---|
| Scan Context [8] | .959 | .788 | .950 | .999 | .642 |
| Scan Context++ [9] | .984 | .850 | .974 | **1.000** | .857 |
| SC+CCD | .981 | .446 | .992 | .996 | .665 |
| RING++ [20] | .995 | .939 | .967 | .998 | .802 |
| Contour Context [7] | .998 | .899 | .995 | **1.000** | .841 |
| OrthoMark, verify top-1 | .999 | .992 | .992 | **1.000** | .950 |
| OrthoMark, verify top-10 | **1.000** | **.995** | **.996** | **1.000** | **.962** |
| ***learned, authors' released weights*** | | | | | |
| OverlapTransformer [15] | .978 | .838 | .924$^\dagger$ | .987$^\dagger$ | .531$^\dagger$ |
| BEVPlace++ [14] | .999$^\dagger$ | .950 | **.990** | **1.000** | .964 |
| BEVPlace++, Ri | 1.000$^\dagger$ | **.997** | .988 | **1.000** | **.974** |
| LCDNet [2]$^§$ | **.996** | .962 | .984 | .989 | .889 |
| ***with the 4 m-protocol cut-off (every method with a pose)*** | | | | | |
| RING++ | .992 | .922 | .980 | .999 | .810 |
| Contour Context | .998 | .892 | .994 | **1.000** | .841 |
| BEVPlace++, Ri | 1.000$^\dagger$ | .996 | .990 | **1.000** | .977 |
| OrthoMark, verify top-10 | **1.000** | **.997** | **.998** | **1.000** | **.978** |
| ***with 3D registration (ICP fitness as the score)*** | | | | | |
| Scan Context + ICP | .987 | .872 | .993 | **1.000** | **.975** |
| RING++ + ICP | **.999** | .974 | **.997** | **1.000** | .915 |
| OrthoMark (top-1) + ICP | **.999** | **.989** | .995 | **1.000** | .948 |

Table 1. KITTI loop closure, LCDNet protocol. **Ri: RANSAC inliers** as the score. OrthoMark verifies within a session with the shared-overlap score and the nearest anchor. Bold / underline: best / second best per block (models evaluated on their training sequence$^\dagger$ excluded). $^§$Released LCDNet weights.

| | AP | | pose-correct AP | | cross-session |
|---|---|---|---|---|---|
| | 01-08 | 02-04 | 01-08 | 02-04 | loop recall |
| Scan Context [8] | .688 | .690 | no pose | | |
| Scan Context++ [9] | .864 | .827 | no pose | | |
| SC+CCD | .831 | .866 | .878 | .900 | .82 |
| RING++ [20] | .726 | .586 | .838 | .765 | .94 |
| Contour Context [7] | .767 | .685 | .835 | .788 | .72 |
| OrthoMark, verify top-1 | .941 | .916 | .979 | .967 | .97 |
| OrthoMark, verify top-10 | **.984** | **.978** | **.986** | **.979** | **.98** |
| ***learned, authors' released weights (no retraining)*** | | | | | |
| OverlapTransformer [15] | .055 | .045 | no pose | | |
| BEVPlace++ [14] | .892 | .889$^‡$ | .905 | .910$^‡$ | **.95** |
| BEVPlace++, Ri | **.941** | **.918**$^‡$ | **.959** | **.953**$^‡$ | **.95** |
| LCDNet [2]$^§$ | .338 | .374 | .154 | .153 | .45$^*$ |
| ***with the 4 m-protocol cut-off*** | | | | | |
| SC+CCD | .899 | .904 | .936 | .935 | |
| RING++ | .807 | .703 | .876 | .814 | |
| Contour Context | .779 | .698 | .840 | .789 | |
| BEVPlace++, Ri | .965 | .943$^‡$ | .982 | .975$^‡$ | |
| OrthoMark, verify top-10 | **.992** | **.982** | **.993** | **.984** | |
| ***with ICP (ICP fitness as the score)*** | | | | | |
| RING++ + ICP | .859 | .734 | .947 | .879 | .95 |
| Contour Context + ICP | .793 | .715 | .851 | .805 | .73 |
| OrthoMark (top-10) + ICP | **.950** | **.874** | **.955** | **.880** | **.98** |

Table 2. NCLT dataset. **Ri: RANSAC inliers**. AP: LCDNet protocol (top-1 within 4 m). Pose-correct AP: relative pose within 2 m / 5°, credited on revisit queries. Cross-session: two robots (9544 closable queries). $^§$Released weights, trained on KITTI. $^*$Estimated on 300 random closable queries (95 % interval ±0.06). $^‡$Checkpoint selected on NCLT sessions including 2012-02-04.

### 4.2. Generalization to new sequences and sensors

All components and parameters were fixed on the sequences of Tab. 1 before these sequences were evaluated. On NCLT, OrthoMark is the best method under every metric: AP 0.984 / 0.978 (0.992 / 0.982 with the cut-off) and pose-correct AP 0.986 / 0.979 within the sessions, ahead of the learned BEVPlace++ (0.941 / 0.918), whose checkpoint was selected on NCLT sessions. Across sessions it closes 0.98 of the loops (BEVPlace++ 0.95) and keeps the highest recall at 100 % precision (0.41; BEVPlace++ 0.25, SC+CCD 0.22, RING++ 0.001), with a 17× smaller descriptor, on one CPU core and 4–38× faster per query (Sec. 4.7). The learned OverlapTransformer and LCDNet do not transfer to NCLT's sensor (AP 0.05 and 0.34–0.37).

### 4.3. Multi-robot loop closure

Tab. 3 shows that OrthoMark matches or exceeds RING++ on all nine sequences, at 1/350 of its descriptor size and without GPU, with the largest gains on the open-field Playground 1–2 and on Square 2, where robots pass places with lateral offsets and in opposite directions. Swarm-SLAM's front end, whose Scan Context descriptor is not translation-invariant, loses 37–56 % of these loops on Square 1–2 and Playground 2. The learned BEVPlace++ retrieves the right place often (R@1 0.77–0.98; Playground 2: 0.37) but, with its KITTI weights on 16-beam scans seen from other viewpoints, estimates a correct pose for fewer of them, so OrthoMark's loop recall is higher on all nine sequences. LCDNet's verification would need about 4 MB of point features per candidate. On Square 1, Square 2 and Library 1 the references themselves contain yaw and inter-robot position errors that lower every method's recall and precision; with a near-miss rule applied equally to all methods, OrthoMark's 150 most confident loops contain no wrong place on Square 1 and 3 on Library 1 (supplementary).

### 4.4. Ablations

The vertical link helps only where it was designed to ( Tab. S1). In verification, replacing the 16 height bits by one occupancy bit, with the same candidates and poses, loses 4 points of loop recall on the cross-viewpoint sequences and 0.043 AP on KITTI 08. In retrieval, putting height layers into the key lowers R@1 by 0.08–0.16 and KITTI 08 AP from 0.82 to 0.37. The 8-bit key and verification from the from compact payload change loops and AP by about 1 % or less (Sec. S1.1).

| loop recall | GrAco | Sq-1 | Sq-2 | Sq-3 | Pg-1 | Pg-2 | Pg-3 | Lib-1 | Lib-2 | KB / keyframe |
|---|---|---|---|---|---|---|---|---|---|---|
| SC+TEASER (Swarm-SLAM) | 0.95 | 0.46 | 0.36 | 0.91 | 0.75 | 0.29 | 0.77 | 0.85 | 0.99 | 9.6 + 38–69 |
| SC+CCD | 0.95 | 0.50 | 0.45 | 0.88 | 0.72 | 0.21 | 0.82 | 0.85 | 0.99 | 9.6 + 0.3–0.6 |
| Contour Context [7] | 0.76 | 0.26 | 0.20 | 0.65 | 0.28 | 0.03 | 0.32 | 0.52 | 0.70 | 24–43 |
| RING++ [20] (GPU) | **1.00** | 0.73 | 0.53 | 0.99 | 0.91 | 0.54 | **0.99** | **0.90** | **1.00** | 675 |
| BEVPlace++ [14] (GPU) | 0.97 | 0.43 | 0.41 | 0.86 | 0.66 | 0.14 | 0.83 | 0.87 | **1.00** | 32 + BEV |
| LCDNet [2] (GPU)* | 0.18 | 0.03 | 0.02 | 0.08 | 0.04 | 0.02 | 0.19 | 0.13 | 0.24 | 1 + ∼4200 |
| OrthoMark, verify top-1 | **1.00** | **0.74** | **0.57** | **1.00** | **0.96** | 0.57 | 0.98 | **0.90** | **1.00** | **1.9 + 1.6–5.3** |
| OrthoMark, top-10 + consensus | **1.00** | **0.74** | **0.57** | **1.00** | **0.96** | **0.66** | **0.99** | **0.90** | **1.00** | **1.9 + 1.6–5.3** |

Table 3. Multi-robot loop recall: share of closable queries whose top-1 has a pose within 2 m / 5° (no ICP). Last column: data exchanged per keyframe (descriptor + verification data per candidate). BEVPlace++: released KITTI (HDL-64) weights on these VLP-16 scans, scored by its RANSAC inliers. *LCDNet: released KITTI-360 weights, posed by the authors' RANSAC on a random sample of 300 closable queries per sequence (all on Sq-1 and Sq-2; 95 % interval within ±0.05). STD, BTC and OverlapTransformer: supplementary.

### 4.5. Mixed sensors

Heterogeneous teams combine sparse and dense lidars. We emulate a 16-beam sensor by keeping 16 of the 64 rings of KITTI (Tab. 4). OrthoMark is the most robust in every case;

| AP | | 64→64 | 16→64 | 16→16 |
|---|---|---|---|---|
| KITTI 00 | Scan Context | .959 | .947 | .952 |
| | SC+CCD | .981 | .750 | .951 |
| | OrthoMark | **.999** | **.982** | **.996** |
| KITTI 08 | Scan Context | .642 | .478 | .474 |
| | SC+CCD | .665 | .302 | .628 |
| | OrthoMark | **.950** | **.727** | **.849** |

Table 4. Mixed sensors: 16-beam queries against a 64-beam map (16 of KITTI's 64 rings); verify top-1. Shaded: ours.

on 00 its top-1 still has a correct pose for all 790 revisits with 16-beam queries. The mismatch costs more than the sparsity itself (08: 0.727 for 16→64 *vs*. 0.849 for 16→16), because the symmetric score penalizes the dense scan for what the sparse sensor never sees. Scoring by how much of the sparser scan the denser one explains raises 08 16→64 (with $S$: 0.599 → 0.749), but lowers same-sensor pairs, so the mode should follow the robots' known sensor types.

### 4.6. Multi-robot SLAM end to end

We replace Swarm-SLAM's lidar front end by OrthoMark and run three robots live on GrAco and on S3E Square 1

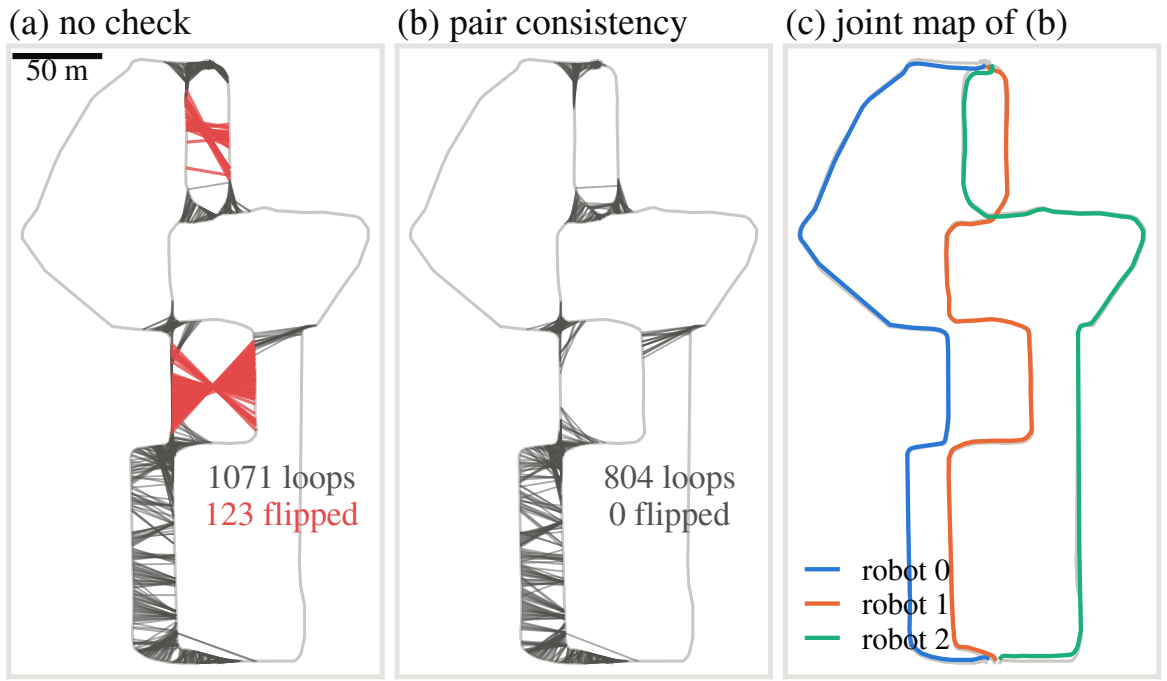

Figure 5. OrthoMark broadcast in Swarm-SLAM on S3E Square 1, inter-robot loops drawn between the reference positions of their two keyframes. (a) Without the robot-pair consistency check, 123 of the 1071 loops with a reference (red) match places across the symmetric square with a yaw error of 173–180°; they come in runs, and the map collapses. (b) With the check, none remains. (c) The joint map of (b), aligned to the reference (gray) in 2D.

(Tab. 5). In Swarm-SLAM's own request-based flow, OrthoMark finds twice the inter-robot loops on GrAco, all correct, and connects the robots on Square 1, where Scan Context passes Swarm-SLAM's threshold across robots only twice (one loop accepted, none in a final graph) and stock never builds a joint map; it uses 5–7× less front-end CPU and 3× less communication. That flow, however, admits a candidate only if its key similarity alone passes a strict threshold and verifies at most one candidate per 5 s, which lets 12 of OrthoMark's loops through on Square 1. Broadcasting key and payload removes this funnel: 835 loops on GrAco with 14.4 MB of communication (stock: 82.2 MB), and 1238 on Square 1 — where the map collapses (127 m), because 11 % of the loops match places across the symmetric square with poses rotated by 180°, in runs of neighbouring keyframes that the consensus and the robust back end accept. The robot-pair consistency gate (Sec. 3.4) removes all of them (Fig. 5) and gives the best Square 1 map of all variants: 980 loops, 2.01 m joint ATE, and every robot below the error of its own odometry (1.7 / 1.7 / 2.2 *vs.* 1.9 / 4.5 / 3.3 m). Of these 980 loops, 96 % agree with the reference positions; the remaining 4 % lie where the S3E reference itself is inconsistent by about 3 m between two robots, and 3D ICP started from them converges with the highest fitness of all loops (supplementary). On GrACo all variants reach about the same ATE (3.0–3.2 m): with planar odometry, even ground-truth loops at the same pairs do not lower it, so the odometry between the loops sets it. With Swarm-SLAM's default 3D odometry, whose roll and pitch drift by up to 51°, the 3D ATE measures that drift rather than the front end (supplementary); all runs use planar odometry.

| | GrACo (3 robots) | | | | S3E Square 1 (3 robots) | | | |
|---|---|---|---|---|---|---|---|---|
| | stock | OrthoMark | | | stock | OrthoMark | | |
| | | request | broadcast | + gate | | request | broadcast | + gate |
| inter-robot loops | 153 | 330 | 835 | **841** | 0 | 12 | 1238 | **980** |
| joint map | yes | yes | yes | yes | failed | yes | collapsed | yes |
| joint ATE [m] | **2.98** | **3.13** | **3.10** | **3.24** | failed | 6.94 | 127.5 | **2.01** |
| ATE per robot [m] | 1.2/3.5/1.7 | 1.1/3.5/1.7 | 1.1/3.5/2.1 | 1.0/3.4/1.8 | 1.8/4.6/2.7† | 1.8/4.6/3.3 | collapsed | **1.7/1.7/2.2** |
| front-end CPU [s] | 7614 | **1357** | 1577 | 1598 | 6366 | **902** | 1603 | 1593 |
| communication [MB] | 82.2 | 28.9 | **14.4** | 14.7 | n/r‡ | **8.5** | 16.0 | 21.2 |
| verification [ms] | 355 | 17 | 61* | 62* | 303 | 10 | 61* | 62* |

Table 5. On GrACo: re-solved offline with ground-truth measurements at the same loop pairs, each run's graph does not improve (2D: 3.01–3.23 m vs. 2.93–3.07 m as run); the differences between the columns come from the odometry of the separate live runs (odometry alone: 3.15–3.89 m), so the four are bolded as equal. Failed: the stock front end never merges the robots; collapsed: wrong loops pull the map apart, so no per-robot error is meaningful. ‡Not reported: Swarm-SLAM's traffic counter records no front-end traffic on this run. Verification time per loop, *per query (top-10 verified).

## 4.7. Runtime and size

Fig. 6 summarizes the trade-off. OrthoMark describes a keyframe in 2.8 ms (VLP-16) to 3.9 ms (HDL-64); retrieval costs about 2 $\mu$s per database keyframe (1.6 ms against 500, 8 ms against 4500), and each verified hypothesis about 3–4 ms, so a query takes 9–16 ms when verifying the top-1 and 60–70 ms for the top-10, on one CPU core. RING++ takes 163–168 ms per description and 48–72 ms per query on the GPU, BEVPlace++ 123–147 ms per description and about 310 ms per query including its pose, and Swarm-SLAM's front end 180–880 ms per query including TEASER++. The key is 1.9 KB (RING++: 675 KB; BEVPlace++: 32 KB; OverlapTransformer: 1 KB without a pose; Scan Context in Swarm-SLAM: 9.6 KB as 1200 float64 values).

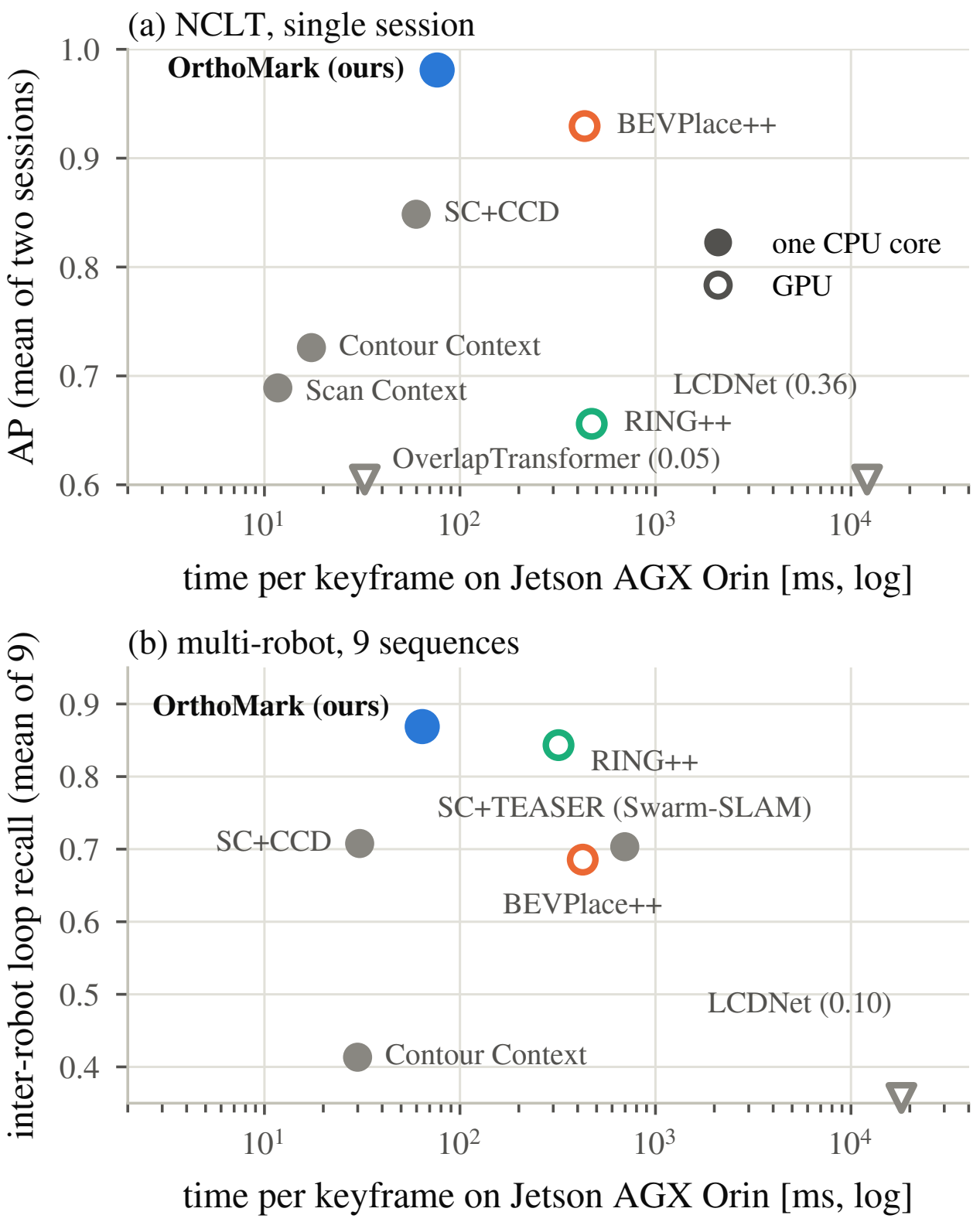

Figure 6. Accuracy against time per keyframe. (a) NCLT, single-session AP averaged over the two sessions. (b) Inter-robot loop recall averaged over the nine multi-robot sequences.

## 5. Conclusion

Height is decisive when two scans are compared cell by cell after alignment, but misleading inside a viewpoint-invariant key. OrthoMark follows this split: an invariant BEV spectrum retrieves and point-anchored height columns verify. Without training, it leads the training-free methods on KITTI, is the best method on NCLT within and across sessions, exceeds every other method on all multi-robot sequences, with fraction of computation and communication.

## Acknowledgments

This work was funded by the European Commission's Marie Skłodowska-Curie Action (MSCA) Project RAICAM (GA101072634).